\documentclass{article}
\usepackage[letterpaper,top=2cm,bottom=2cm,left=3cm,right=3cm,marginparwidth=2cm]{geometry}
\usepackage{graphicx} 
\usepackage{biblatex}
\usepackage[colorlinks=true, allcolors=blue, breaklinks=true]{hyperref}
\usepackage{multicol}
\title{Open the Data! Chuvash Datasets}
\author{Nikolay Plotnikov, Alexander Antonov\footnote{Corresponding author \href{https://github.com/AlAntonov}{https://github.com/AlAntonov}}}
\date{June 2024}

\begin{document}

\maketitle

\begin{multicols}{2}

\begin{abstract}
In this paper, we introduce four comprehensive datasets for the Chuvash language, aiming to support and enhance linguistic research and technological development for this underrepresented language. These datasets include a monolingual dataset, a parallel dataset with Russian, a parallel dataset with English, and an audio dataset.

Each dataset is meticulously curated to serve various applications such as machine translation, linguistic analysis, and speech recognition, providing valuable resources for scholars and developers working with the Chuvash language. Together, these datasets represent a significant step towards preserving and promoting the Chuvash language in the digital age.
\end{abstract}

\section{Introduction}

In recent years, the development of artificial intelligence (AI) has made remarkable strides, significantly transforming various fields. This progress is not limited to English but extends to numerous other languages, enhancing accessibility and fostering innovation across diverse linguistic landscapes.

Collecting dataset and making them publicly available under an open license is crucial for advancing research and development in natural language processing.

In this paper we present 4 datasets in the Chuvash language.

\section{Chuvash Monolingual}

The first dataset we present is the Chuvash monolingual dataset. This resource is invaluable for a wide range of natural language processing (NLP) tasks, including training large language models and developing machine translation systems. This dataset, along with the others, is readily accessible via the link on Hugging Face\footnote{\href{https://huggingface.co/datasets/alexantonov/chuvash_mono}{huggingface.co/datasets/alexantonov/chuvash\textunderscore mono}}

Dataset size is 3.9 million sentences.

\section{Chuvash-Russian Parallel}

The second dataset is a primary parallel dataset. Given the close relationship between Chuvash and Russian, we have compiled a Chuvash-Russian parallel dataset\footnote{\href{https://huggingface.co/datasets/alexantonov/chuvash_russian_parallel}{huggingface.co/datasets/alexantonov/chuvash\textunderscore russian\textunderscore parallel}}.

This dataset features roughly 1.4 million sentence pairs, primarily collected manually, to ensure quality and accuracy.

\section{Chuvash-English Parallel}

The Chuvash-English parallel corpus\footnote{\href{https://huggingface.co/datasets/alexantonov/chuvash_english_parallel}{huggingface.co/datasets/alexantonov/chuvash\textunderscore english\textunderscore parallel}} was created automatically from the previous dataset and comprises 200 thousand sentence pairs sourced from books.

The main goal of the dataset is to add the Chuvash language to English-centric models. In other cases, it is more useful to use the Chuvash-Russian Parallel dataset

\section{Chuvash Voice}

The fourth dataset is an audio corpus of the Chuvash language\footnote{\href{https://huggingface.co/datasets/alexantonov/chuvash_voice}{huggingface.co/datasets/alexantonov/chuvash\textunderscore voice}}. It is designed to address tasks related to automatic speech recognition (ASR) and speech synthesis (TTS). In the future, this corpus will also be valuable for developing multimodal large language models for Chuvash.

It is important to note that the dataset was collected by analogy with another well-known Common Voice corpus\cite{ardila2020common}, which is the de facto standard in the industry. Therefore, both corpora can be used together to train models, since they are easy to combine. An example code is shown on the dataset page.

Initially, audio files were collected in different formats: mp3, webm, ogg. For convenience, we converted all files into the same format, we chose mp3. Information about the original format is stored in the name of the audio file.

The dataset contains about 30 thousand records or 38 hours. Together with Common Voice, which contains 25 hours in the Chuvash language, we have 63 hours of recordings.

The dataset contains a large volume of files recorded by one unique voice, more precisely 30 hours. This means it can be used for TTS task.

\printbibliography 

@misc{ardila2020common,
      title={Common Voice: A Massively-Multilingual Speech Corpus}, 
      author={Rosana Ardila and Megan Branson and Kelly Davis and Michael Henretty and Michael Kohler and Josh Meyer and Reuben Morais and Lindsay Saunders and Francis M. Tyers and Gregor Weber},
      year={2020},
      eprint={1912.06670},
      archivePrefix={arXiv},
      primaryClass={cs.CL}
}

\end{multicols}

\end{document}